\documentclass[10pt, a4paper]{article}

\usepackage[final]{lrec2026} %

\usepackage{comment} %
\usepackage{lingex}
\usepackage{tikz-dependency}

\newcommand{\udpdtcz}[0]{UD\_Czech-PDTC}
\newcommand{\udpdtc}[0]{\textbf{\udpdtcz{}}}

\newcommand{\exee}[1]{`{#1}'}

\newcommand{\deptrans}[1]{\node (t) at (\matrixref.south) [yshift=-1mm] {\exee{#1}};}

\usepackage{booktabs} %
\usepackage{multirow} %
\usepackage{pifont} %
\newcommand{\YES}{\ding{51}} %
\newcommand{\NO}{{\color{gray!50}\ding{55}}} %

\tikzstyle{word}=[draw=blue!80!black, shade, top color=blue!40, rounded
corners]
\newcommand{\udscale}{0.9}

\title{Meet \udpdtcz{}: \\ A Large and Genre-Rich Treebank in Universal Dependencies}

\name{Marie Mikulová, Barbora Štěpánková, Daniel Zeman, Jan Štěpánek,  \\
 \large\textbf{Milan Straka, Jan Hajič }}

\address{Charles University, Faculty of Mathematics and Physics \\
         Institute of Formal and Applied Linguistics \\
         Malostranské náměstí 25, Prague, Czechia \\
         \{mikulova,stepankova,zeman,stepanek,straka,hajic\}@ufal.mff.cuni.cz\\}

\abstract{
Czech has been part of Universal Dependencies since its first release in 2015.
It has also been one of the best represented languages, with the Prague Dependency Treebank
being order of magnitude larger than most other UD treebanks.
More recently, three other datasets from the Prague family were added and the annotations
thoroughly revisited, forming the ``Prague Dependency Treebank-Consolidated'' (PDT-C).
In comparison to the original PDT, PDT-C is more than twice as large, but it is
also much more diverse in terms of genres and domains.
In this paper, we describe the conversion of the new resource to Universal Dependencies.
While the two annotation schemes are relatively similar at the first sight, there
are numerous small differences in topology of the dependency structures and in
granularity of the POS and relation type inventories.
We demonstrate a selection of such differences on examples, discuss the diverging
motivations, as well as ways to overcome the differences during conversion.
We argue that while PDT is less ``universal'' and more tightly bound to one language,
its multi-layer annotation is rich and provides all information needed for basic
UD trees, and much more.
\\ \newline
}

\usepackage{fancyhdr}
\fancypagestyle{officialbibref}{\fancyhf{}\fancyheadoffset[L,R]{50pt}\fancyhead[C]{\kern-5em This paper was published at \textbf{LREC 2026} -- please cite the published version {\small\url{https://aclanthology.org/2026.lrec-1.909/}}.\kern-5em}\fancyfoot[C]{\normalsize\thepage}\addtolength{\topmargin}{-50pt}\addtolength{\headsep}{50pt}}
\begin{document}
\thispagestyle{officialbibref}
\pagenumbering{arabic}\pagestyle{plain}

\maketitleabstract

\section{Introduction}
\label{introduction}

An important application and promotion of an annotated corpus is its
conversion into another formalism. The significance of such conversion has
several aspects: (i) it can highlight similarities and differences in the
frameworks, (ii) emphasize typological differences among languages in terms
of their ability to incorporate or exclude specific linguistic phenomena, and
(iii) integration into various concepts demonstrates the robustness and
universality of the chosen format.

In our contribution, we aim to demonstrate
the richness of linguistic annotation present in the Prague Dependency
Treebank \cite{pdtc-2026}, linking morphology, syntax, and semantics.
This is evidenced, among other things, by the fact that the treebank is used
for conversions into various frameworks (e.g., Uniform Meaning Representation
\cite{itatumr2024,stepanek-etal-2025-comparing}, Minimal Recursion Semantics
\cite{jakob-etal-2010-mapping}, CorefUD \cite{nedoluzhko-etal-2022-corefud},
Meaning Representation Parsing format used in CoNLL shared tasks
\cite{oepen-etal-2019-mrp,oepen-etal-2020-mrp}, the Penn Discourse Treebank
format \cite{MiSyPresentingPDiT40-2026}). Here, we focus on its conversion to Universal
Dependencies %
\cite{de-marneffe-etal-2021-universal} and elaborate the aspects (i)--(iii)
listed  above.

\begin{figure}[t!]
  \begin{center}
  \includegraphics[scale=0.35]{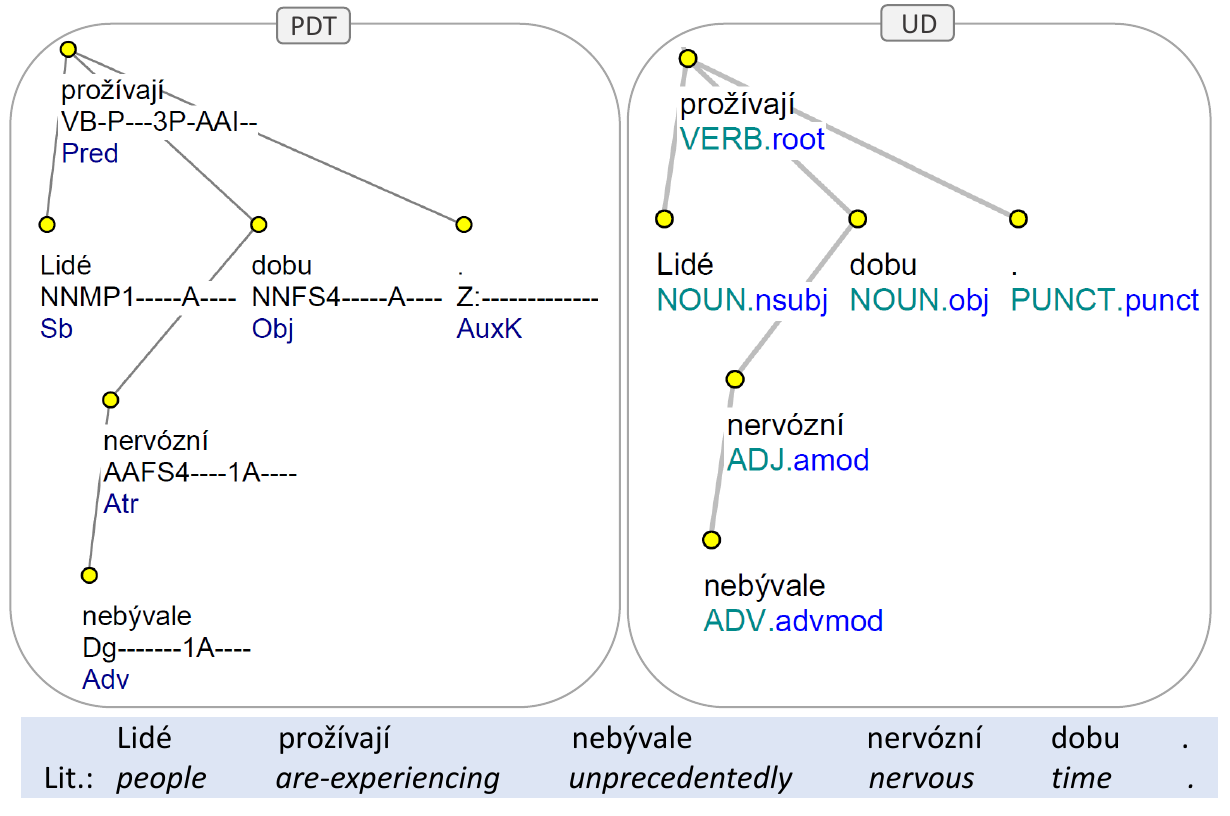}
  \caption{PDT sentence representation and its conversion to UD.}
  \label{fig:UD-PDT-same}
  \end{center}
\end{figure}

The \udpdtc{} treebank (available first in UD version 2.16;
\citealplanguageresource{ud216}) is a UD conversion of the Prague Dependency
Treebank -- Consolidated, released in 2024 (PDT-C 2.0;
\citealplanguageresource{pdtc20}), and with 3440K words is now one of the
largest treebanks in UD, containing texts of various genres. In addition to
the pilot Prague Dependency Treebank of journalistic texts (enriched version
from 2006; \citealplanguageresource{pdt20}, converted to the first UD 1.0
version in 2015; \citealplanguageresource{UDv12015}), the consolidated PDT-C
release includes three other PDT-corpora of spoken, translated, and
user-generated content, all of which have undergone significant changes in
surface syntax annotation, supplemented by complete manual annotation, using
a uniform annotation scheme.

\begin{figure*}[ht!]
  \begin{center}
    \scalebox{\udscale}{%
    \begin{dependency}
      \begin{deptext}[row sep=.1ex]
        VERB            \& PRON   \& DET \& PART    \& ADP \& NOUN     \& PUNCT \\
        |[word]| Vysvětlíme \& |[word]| vám \& |[word]| to \& |[word]| snad \& |[word]| na \& |[word]| příkladu \& |[word]| . \\
        we-will-explain \& to-you \& it  \& perhaps \& on  \& example  \& .     \\
      \end{deptext}
      \depedge[edge unit distance=2.5ex]{1}{7}{punct}
      \depedge[edge unit distance=2.4ex]{1}{6}{obl}
      \depedge{1}{4}{advmod}
      \depedge{1}{3}{obj}
      \depedge{1}{2}{obl:arg}
      \depedge{6}{5}{case}
      \deproot[edge unit distance=4.5ex]{1}{root}
      \deptrans{Let us explain this to you with an example.}
    \end{dependency}%
    }
    \includegraphics[scale=0.6]{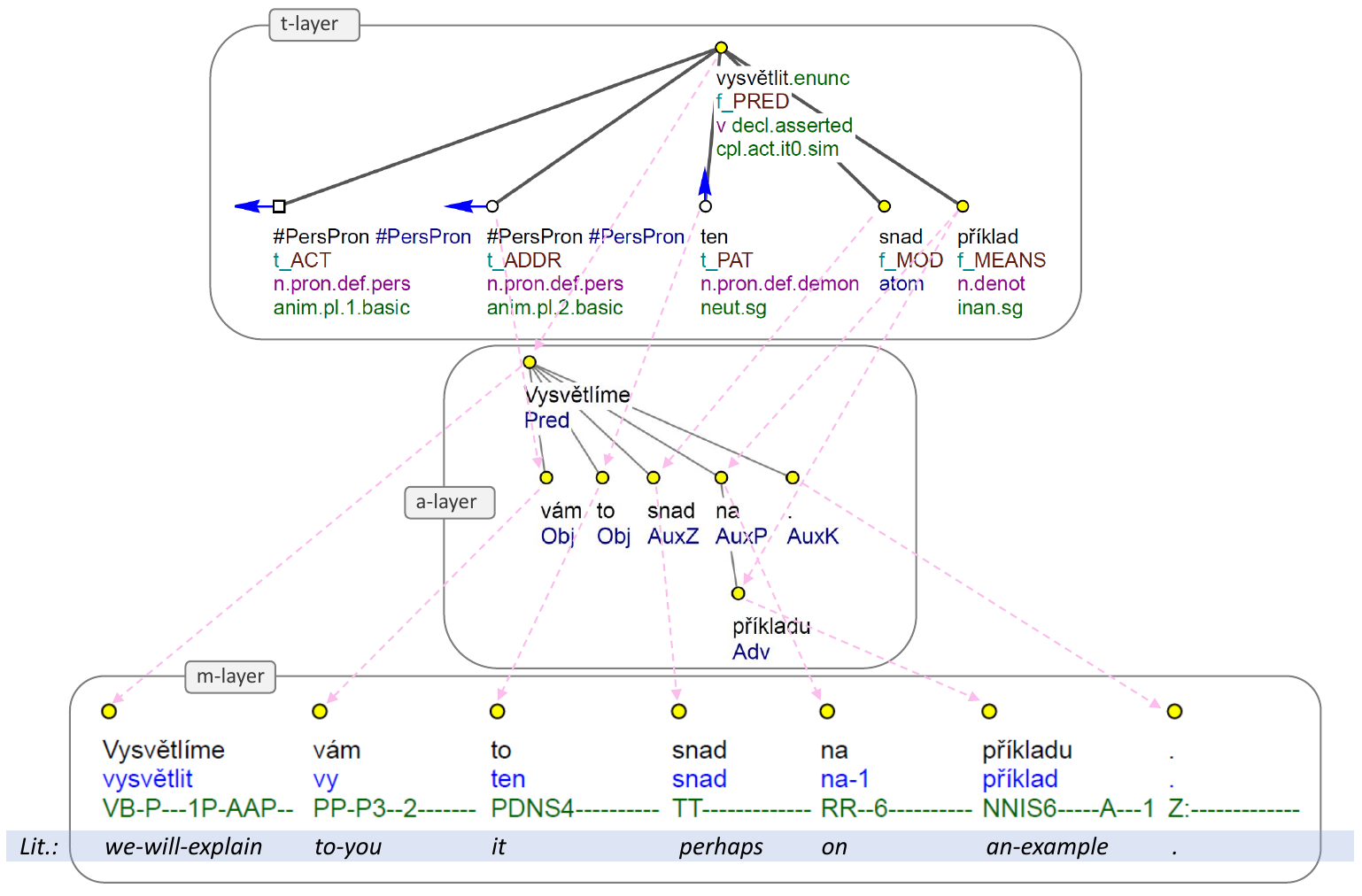}
    \caption{Representation of the Czech sentence (\ref{ex:main}) in UD (top) and PDT (bottom) annotation scheme.}
    \label{fig:UD-PDT}
  \end{center}
\end{figure*}

In the contribution, we discuss the major differences between the two formalisms (theoretical foundations, approaches to syntax, morphology, and
semantics). Using specific examples from various areas of language
description (part-of-speech classification, coordination, ellipsis,
semantic-pragmatic expressions), we discuss how these differences can
influence data interpretation or linguistic research (the loss of specific
phenomena, the adaptation of the format to new phenomena). Most importantly,
we highlight that the morphosyntactic annotation in the PDT and UD frameworks
is highly similar, as illustrated in Fig.~\ref{fig:UD-PDT-same}.

The paper is organized as follows: the following subsections provide a brief
description of both the frameworks (UD in~\ref{ud}, PDT in~\ref{pdt}), the converted
language (\ref{czech}), and history of PDT-to-UD conversions (\ref{history}). The
conversion is the focus of Sect.~\ref{conversion}. We discuss conversion of
part-of-speech taxonomy (\ref{pos}), syntactic structure (\ref{structure}), and syntactic relations
(\ref{relation}). Sect.~\ref{enhanced} addresses the Enhanced UD. The
conversion is summarized in Sect.~\ref{discussion}. The Czech UDPipe model is
presented in Sect.~\ref{udpipe} and Sect.~\ref{conclusion} concludes the paper.

\subsection{Universal Dependencies}
\label{ud}

Universal dependencies (UD; \citealp{de-marneffe-etal-2021-universal}) is a
stunning project -- framework for consistent morphosyntactic annotation
across different languages. It is an open community effort, now with more
than 700 contributors who created 319 treebanks in 179 languages. The general
philosophy is to provide a universal inventory of categories and guidelines
and to facilitate consistent annotation of similar constructions across
languages, while allowing language-specific extensions when necessary. This
effort is a good basis for cross-linguistically consistent annotation of
typologically diverse languages in a way that supports computational natural
language understanding as well as broader linguistic studies. The detailed
specification of UD annotation is available on the project
website\footnote{\url{https://universaldependencies.org/}} and the overview
can be also found in \citet{nivre-etal-2020-universal} or
\citet{de-marneffe-etal-2021-universal}. Fig.~\ref{fig:UD-PDT} demonstrates
the UD annotation scheme on the Czech example (\ref{ex:main}).

\begin{exe}
  \ex \label{ex:main}{\small \textit{Vysvětlíme vám to snad na příkladu.} \\
  we-will-explain to-you it perhaps on an-example. \\
  ‘Let us explain this to you with an example.’}
\end{exe}

The UD framework is based on a lexicalist view of syntax, which means that
dependency relations hold between words, and that morphological features are
encoded as properties of words with no attempt at segmenting words into
morphemes. The \textbf{morphological specification} of a word in the UD
scheme consists of three types of information: a lemma representing the base
form of the word, a part-of-speech (POS) tag representing the grammatical
category of the word, a set of features representing lexical and grammatical
properties associated with the particular word form. For simplicity, only POS
annotation is shown in Fig.~\ref{fig:UD-PDT}.

\textbf{Syntactic annotation} consists of labeled dependency relations
between words. The basic representation forms a tree rooted in one word,
normally the main sentence predicate (the verb \textit{vysvětlíme}
‘we-will-explain’ in Fig.~\ref{fig:UD-PDT}), on which other words of the
sentence (including punctuation) depend. The syntactic analysis gives
priority to predicate-argument and modifier relations that hold directly
between content words, as opposed to being mediated by function words, so all
function words including copula are uniformly treated as dependents (cf.
treatment of prepositional phrase \textit{na příkladu} ‘on example’ in
Fig.~\ref{fig:UD-PDT}).

In addition to the basic representation, which is obligatory for all UD
treebanks, it is possible to provide an \textbf{enhanced representation}
\cite{nivre-etal-2020-universal}, which adds (and in a few cases changes)
relations in order to give a more complete basis for semantic interpretation
(details in Sect.~\ref{enhanced}).

\subsection{Prague Dependency Treebank}
\label{pdt}

The long-run project of Prague Dependency Treebank (PDT; \citealp{pdtc-2026}) is
unique in its attempt to systematically cover and link different layers of
language description including a rich semantic representation. Its
hierarchical multi-layer architecture is based on the theory of Functional
Generative Description \cite{meaningSgall1986} and is described in several
detailed annotation manuals available from the project web
site.\footnote{\url{https://ufal.mff.cuni.cz/pdt-c}}

The PDT annotation scheme is illustrated in Fig.~\ref{fig:UD-PDT} on the same
Czech example (\ref{ex:main}) as we used for UD. In Fig.~\ref{fig:UD-PDT},
each annotation layer of the PDT-system is displayed in a separate box. The
links between the layers are indicated by the pink dashed arrows. The
original raw text is stored at the lowest layer of the system (and it is not
shown in Fig.~\ref{fig:UD-PDT}).
Above the raw text layer, there are three layers of annotations:
morphological, surface syntactic, and deep syntactic.

Czech is a highly inflectional language (see Sect.~\ref{czech}). At the
\textbf{morphological layer} (\textit{m-layer} box in Fig.~\ref{fig:UD-PDT}),
each word form is described by a 15-character tag that specifies its lexical
and grammatical properties. All tokens are also assigned a POS category
within the tag (the first position).

Above the linearly structured m-layer, there are two syntactic layers, the
\textbf{analytical} (\textit{a-layer}) reflecting surface dependency
structure and the \textbf{tectogrammatical} (\textit{t-layer}) reflecting the
deep syntactic structure.  At the a-layer, which is closest to the UD
representation, a syntactic structure is captured by a rooted directed tree
structure with the specification of the head for each word and the assignment
of a label describing dependency syntactic relation (such as subject, object,
adverbial). As in UD, the root of the tree is typically the predicate of a
sentence.

At the a-layer, a distinction is made between different groups of function
words. The key division is between function words that operate within verbal
complexes and encode morphosyntactic properties of verbs (i.e., auxiliaries),
and those that are part of nominal groups (prepositions) or link clauses into
a single unit (conjunctions). Auxiliaries are analyzed as dependents of the
verb they “belong” to, whereas prepositions and conjunctions are treated as
the heads of the nouns or clauses whose form they “govern”; cf. treatment of
\textit{na příkladu} ‘on example’ in Fig.~\ref{fig:UD-PDT}.

The t-layer captures rich semantic annotations of a sentence:
predicate--argument structure, semantic classification of adjuncts, semantic
counterparts of morphological categories, topic--focus articulation,
information structure, coreference, ellipsis, and discourse relations. The
main difference between the two PDT syntactic layers is how words correspond
to nodes: at the a-layer, every word of the raw text (including punctuation)
is represented by a tree node and no additional nodes are allowed (this
approach is the same as in UD), while the t-layer consists of nodes that only
represent content words (with some exceptions such as coordination); function
words such as prepositions, auxiliary verbs, etc. are not present. There is
for example only one node for the prepositional phrase \textit{na příkladu}
‘on example’ at t-layer annotation in Fig.~\ref{fig:UD-PDT}. The contribution
of function words to the meaning of the sentence is captured within the
complex labels of the content word nodes -- see many values attached to the
nodes at the t-layer. At the t-layer, new nodes are also added for semantic
units deleted on the surface; in Fig.~\ref{fig:UD-PDT} the restoration of a
deletion is illustrated by the \texttt{\#PersPron} (personal pronoun) node
for the unexpressed subject ({\tt ACT}) of the sentence predicate.

The latest version, \textbf{Prague Dependency Treebank – Consolidated 2.0}
\citelanguageresource{pdtc20}\footnote{\url{http://hdl.handle.net/11234/1-5813}}
is a manually annotated, genre-diversified, consolidated release of the
existing PDT-corpora of Czech data, uniformly annotated using the PDT scheme
described above. PDT-C includes four datasets: Prague Dependency Treebank
(written newspaper and journal texts); Czech part of Prague Czech-English
Dependency Treebank (business news, translated from English), Prague
Dependency Treebank of Spoken Czech (spoken data); PDT-Faust (user-generated
texts). For the PDT-C 2.0 version, manual annotation at the a-layer is
performed in those parts of the corpus that were previously annotated only by
automatic tools. Unlike previous versions, PDT-C~2.0 provides fully manual
annotation at all three annotation layers in all datasets, which ensures that
its UD conversion has the quality required for official UD releases.

\subsection{Czech}
\label{czech}

Czech is a West Slavic language spoken by ca.\ 10 million people, most of
them living in Czechia. As a national language, it is also one of the
official languages of the European Union. Czech is an inflectional language
characterized by a complex system of nominal declension and verbal
conjugation; its word order is very flexible, driven by pragmatic factors
rather than syntax. It uses a Latin-based alphabet supplemented with
diacritical marks for long vowels and palatalized consonants.
In the field of NLP/CL, Czech cannot be considered a low-resource language,
not least thanks to the Prague Dependency Treebank, which was in fact a
pioneering effort and one of the earliest treebanks ever created. There is
also a wealth of other corpora (esp. \textit{Czech National Corpus}) and
resources available, including morphological analyzers, valency lexicons, or
the Czech \textit{WordNet}.

\subsection{History of PDT to UD Conversion}
\label{history}

The PDT data has been part of UD since the very beginning. The first release
of UD (\textbf{v1.0}; \citealplanguageresource{UDv12015}) in 2015 contained
treebanks of 10 languages, including conversion of Prague Dependency Treebank
2.0 from 2006 \citelanguageresource{pdt20}. In 2017, UD version 2.0
\citelanguageresource{ud20} was published, which differs significantly from
the previous releases. Changes in the guidelines \textbf{from v1 to v2} are
summarized in \citet{nivre-etal-2020-universal}, refering to release 2.5,
covering 90 languages.

The consolidated release of \textbf{PDT-C 2.0}, containing 3440K words,
appears in UD for the first time in version~2.16
\citelanguageresource{ud216}. Together with UD\_German-HDT
\cite{hdt-ud}, they are the largest treebanks in UD.

The first treebank featuring the \textbf{Enhanced UD} representation
was the English Web Treebank \cite{schuster-manning-2016-enhanced} in version
2.2 from 2018.
PDT appeared in the enhanced UD format for the first time in version~2.6
(2020). The most recent conversion of PDT-C 2.0 also includes
enhanced UD relations, which are based on the original manual annotation.
\udpdtc{} is one of the 19 UD treebanks that have all six types of
enhancements defined in the guidelines.

\begin{figure*}[t]
  \begin{center}
    \scalebox{\udscale}{%
    \begin{dependency}
      \begin{deptext}[row sep=.1ex]
        DET     \& NOUN      \& AUX  \&[1mm] ADJ     \& PUNCT \& ADJ    \& PART   \& PUNCT \\
        |[word]| Některá \& |[word]| řešení \& |[word]| jsou \& |[word]| podobná \& |[word]| , \& |[word]| jiná \& |[word]| nikoli \& |[word]| . \\
        Some    \& solutions \& are  \& similar \& ,     \& others \& not    \& .     \\
      \end{deptext}
      \depedge{2}{1}{det}
      \depedge{4}{2}{nsubj}
      \depedge{4}{3}{cop}
      \deproot{4}{root}
      \depedge[edge unit distance=2.25ex]{4}{8}{punct}
      \depedge{4}{6}{conj}
      \depedge{6}{5}{punct}
      \depedge{6}{7}{orphan}
      \deptrans{Some solutions are similar, others are not.}
    \end{dependency}%
    }
    \caption{UD representation of sentence  (\ref{ex:nekterareseni}).}
    \label{fig:nekterareseni-UD}
  \end{center}
\end{figure*}

\begin{figure}[ht]
  \begin{center}
  \includegraphics[scale=0.4]{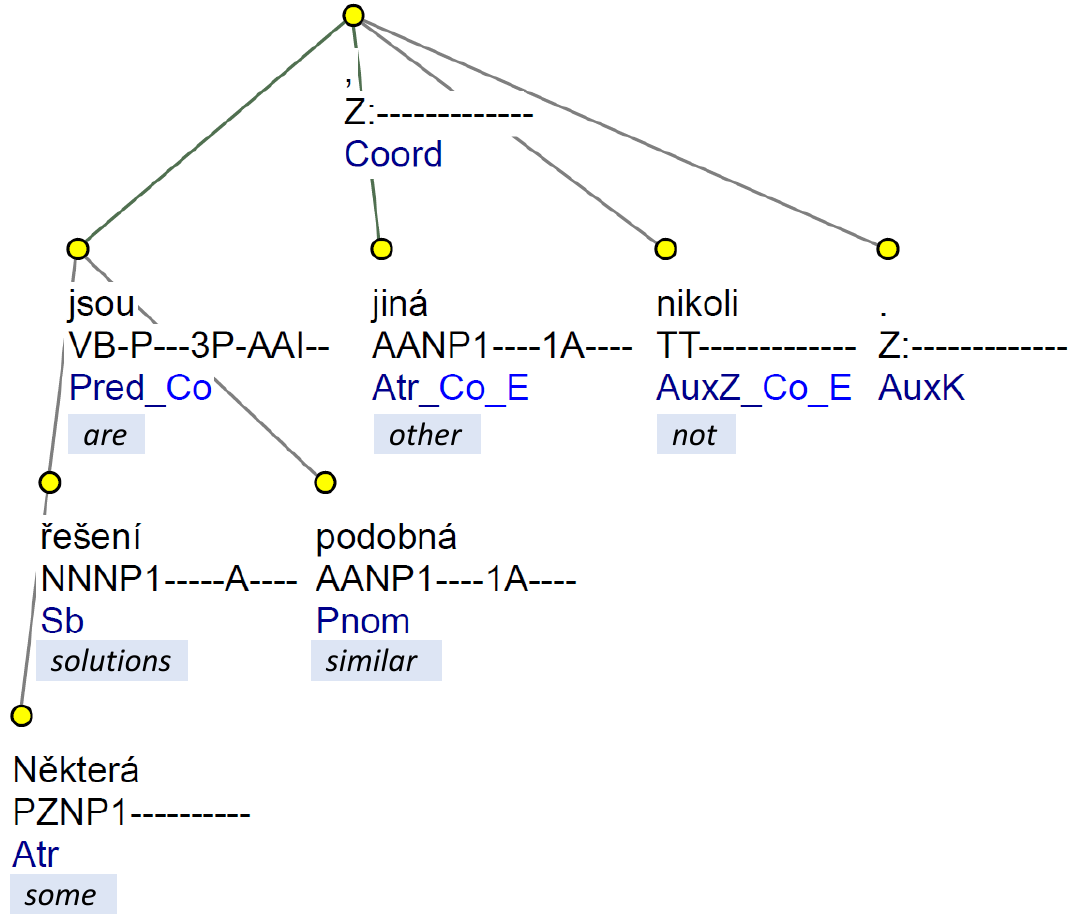}
  \caption{PDT representation of sentence (\ref{ex:nekterareseni}).}
  \label{fig:nekterareseni-A}
  \end{center}
\end{figure}

\section{PDT to UD Conversion}
\label{conversion}

As shown in Fig.~\ref{fig:UD-PDT-same},\footnote{There are various approaches
to visualizing dependency annotations. The standard visualization of PDT is
provided by the TrEd tool (\url{https://ufal.mff.cuni.cz/tred/}). For simplicity, we display m-layer tags in a-layer tree.
UD trees are typically shown as arcs over a linear sentence, which we mostly
follow except for the side-by-side comparison of the two frameworks in
Fig.~\ref{fig:UD-PDT-same}.} the representation of a sentence in PDT and its
counterpart in UD can be nearly identical, disregarding the different naming
conventions for POS tags and syntactic relations. This similarity arises from
the fact that the original UD proposal was heavily influenced by
Indo-European languages and Czech belongs to this family. Both frameworks are
based on dependency syntax, in contrast to, e.g., constituency-based syntax,
which is traditional in American linguistics \cite{chomsky1957syntactic}.
They both treat the predicate as the head of the syntactic structure. In the
basic surface syntactic representation (the a-layer in PDT and basic
representation in UD), every word in the sentence, including punctuation, has
a corresponding node in the tree, and no extra nodes are allowed. The core
inventories of part-of-speech tags and syntactic relations also largely
overlap (cf. POS annotation in Fig.~\ref{fig:UD-PDT-same}: {\tt VERB} in UD
vs. {\tt V} in PDT for verbs; {\tt NOUN} vs. {\tt N} for nouns, {\tt ADJ} vs.
{\tt A} for adjectives, {\tt ADV} vs. {\tt D} for adverbs, {\tt PUNCT} vs.
{\tt Z} for punctuation; and syntactic relations annotation: {\tt root} in UD
vs. {\tt Pred} in PDT for the tree root; {\tt nsubj} vs. {\tt Sb} for
subject; {\tt obj} vs. {\tt Obj} for object, {\tt advmod} vs. {\tt Adv} for
adverbial).

The main difference is that PDT uses a multi-layer scheme separating
linguistic information, while UD integrates all (mainly morphosyntactic)
information into a single graph. In the following sections, we primarily
compare the UD representation and PDT representation at the m-layer (the
basis for POS and features conversion) and the a-layer (the basis for
conversion of syntactic structure and relations).\footnote{Annotation at the
m/a-layer in PDT may in some respects appear less fine-grained than in UD.
However, there is also t-layer and differentiation of some distinctions
introduced in UD is performed up to this layer.} To illustrate the PDT-to-UD
conversion process, we have selected several `clean' cases where the
conversion is clear and consistent, as well as several more complex instances
where pragmatic decisions often need to be made that may result in either
over- or under-interpretation of the original syntactic structure.

\subsection{Part-of-Speech Categories}
\label{pos}

There are 17 universal POS categories in UD, including determiner ({\tt DET})
and auxiliary ({\tt AUX}), which are not part of the Czech/PDT POS
taxonomy.\footnote{There is also {\tt PROPN} category used for proper names. In PDT,
proper names are not addressed within the POS category but rather encoded as
specific information in the lemma of the given word at the m-layer.} During
the conversion, we adopted a strategy of tagging words that satisfy the UD
definitions of determiners and auxiliaries, aiming for universality and
facilitating cross-linguistic comparison.

\subsubsection{Determiners}

The notion of determiners ({\tt DET}) is unknown in traditional Czech
grammars (in Czech, there are no articles as in English). Words equivalent to
English determiners are traditionally classified as pronouns, adjectives
and/or numerals. Nevertheless, traditional pronouns can be divided to
subclasses with distinct morphosyntactic behavior, namely those that resemble
adjectives and those that do not. `Adjective-like pronouns' can modify
nominals and inflect for gender (and also number and case) to express
agreement with the modified noun. Many of them can also stand alone as
nominal heads. However, since UD v2 it is not necessary to distinguish them
solely by context. We can pre-categorize words in the dictionary, and that is
what we do in this case. If a `traditional pronoun' can show gender agreement
with a modified noun, we tag it {\tt DET} regardless of whether it modifies
or heads a nominal (e.g., the demonstrative \textit{to} `the/it' is tagged as
a pronoun ({\tt P}) in PDT, and it has always the {\tt DET} category in UD --
even if it functions as an object as in (\ref{ex:main});  cf.
Fig.~\ref{fig:UD-PDT}).

\subsubsection{Auxiliaries}

According to UD definition, an auxiliary ({\tt AUX}) is a function word that
accompanies the lexical verb and expresses grammatical distinctions not
carried by the lexical verb, such as person, number, tense, mood, aspect,
voice or evidentiality (so-called TAMVE markers). Since UD v2 the {\tt AUX}
category also includes the copula. In Czech, all these functions are
fulfilled by the verb \textit{být} `to be'.

In PDT, all occurrences of \textit{být} ‘to be’ are uniformly tagged as verbs
and the context-based distinction between auxiliary and full verb is left for
the syntactic annotation. In the PDT to UD conversion, we took the side of
pre-categorizing this verb as {\tt AUX} in dictionary, on the grounds that
only a small minority of cases could pass as full verbs by UD guidelines.
Like in the PDT framework, the distinction is reflected in syntactic relations rather than
the POS tag.

In \textbf{copular construction}, PDT treats the copula
\textit{být} ‘to be’ as the head due to the strong agreement principle in
Czech; the nominal predicate is assigned the {\tt Pnom} relation. In UD, the
relation is inverted and labeled {\tt cop}. The different treatment of copula
in PDT and UD is illustrated by the example \textit{řešení jsou podobná}
‘solutions are similar’ (\ref{ex:nekterareseni}) in
Fig.~\ref{fig:nekterareseni-A} and~\ref{fig:nekterareseni-UD}, respectively.

\begin{exe}
  \ex \label{ex:nekterareseni}{\small \textit{Některá řešení jsou podobná, jiná nikoli.} \\
  some solutions are similar, others not. \\
  ‘Some solutions are similar, others are not.’}
\end{exe}

\subsection{Dependency Structure}
\label{structure}

In terms of capturing the dependency structure of a sentence, the main
differences between the PDT and UD frameworks, aside from the treatment of
function words,\footnote{The status of function words in various dependency
frameworks is the subject of a special issue of \textit{Linguistic Analysis} journal. The treatment of function words in the UD framework is described in \citet{LA-UD} and in the
PDT framework in \citet{LA-PDT}.} are coordination and ellipsis.

\subsubsection{Coordination}

\begin{figure}[t!]
  \begin{center}
    \includegraphics[scale=0.22]{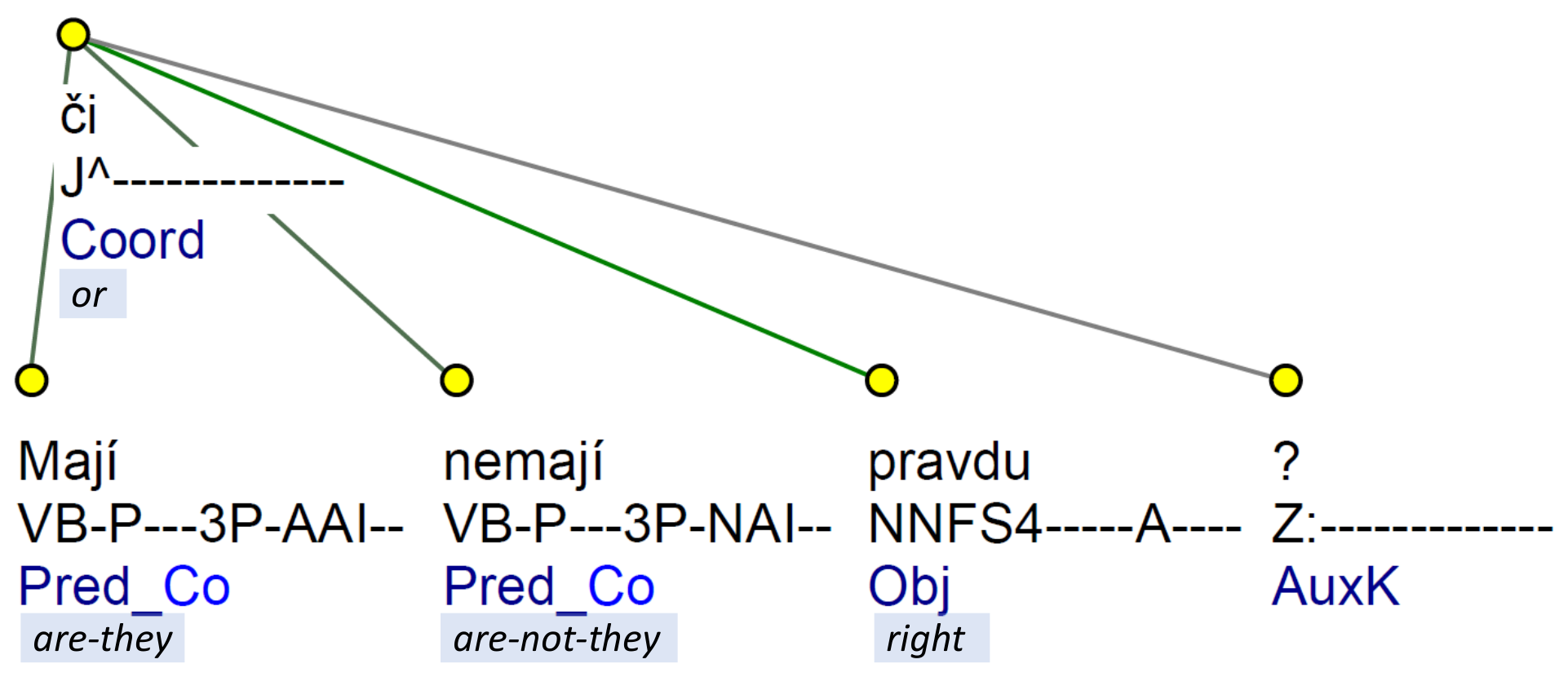}
    \caption{PDT representation of sentence (\ref{ex:majicinemaji}).}
    \label{fig:majicinemajiA}
  \end{center}
\end{figure}

Coordination is a complicated phenomenon in any dependency formalism
\cite{popel-etal-2013-coordination}. In PDT, a coordinating conjunction (cf.\
Fig.~\ref{fig:majicinemajiA}) or a punctuation mark (cf.\
Fig.~\ref{fig:nekterareseni-A}) is made a ``technical head'' of the
coordination and gets the label {\tt Coord} as a marker of the coordination
construction. The conjuncts are formally represented as its dependents and
they each carry a flag (displayed as the {\tt \_Co} suffix in tree diagrams)
to distinguish them from real dependents; cf.\ the PDT representation of
(\ref{ex:majicinemaji}) with two coordinate predicates in
Fig.~\ref{fig:majicinemajiA}.

In UD, there is a different strategy to capture coordination. The first
conjunct is treated as the parent (or “technical head”) of all following
conjuncts via the {\tt conj} relation. Coordinating conjunctions and
punctuation delimiting the conjuncts are attached to their associated conjuncts
using the {\tt cc} and {\tt punct} relations, respectively. See
Fig.~\ref{fig:majicinemajiUD} depict the UD representations of sentence
(\ref{ex:majicinemaji}).

\begin{exe}
  \ex \label{ex:majicinemaji}{\small \textit{Mají či nemají pravdu?} \\
   they-have or they-have-not truth?\\
  ‘Are they right or wrong?’}
\end{exe}

Despite the different representation of coordination, the PDT to UD
conversion is relatively straightforward. %
However, coordination may form very intricate structures when combined with
ellipsis.

\begin{figure}[t!]
  \begin{center}
    \scalebox{\udscale}{%
    \begin{dependency}
      \begin{deptext}[column sep=-1mm, row sep=.1ex]
        VERB      \& CCONJ \& VERB          \& NOUN   \& PUNCT \\
        |[word]| Mají \& |[word]| či \& |[word]| nemají \& |[word]| pravdu \& |[word]| ? \\
        they-have \& or    \& they-have-not \& truth  \& ?     \\
      \end{deptext}
      \depedge{1}{5}{punct}
      \depedge{1}{4}{obj}
      \depedge{1}{3}{conj}
      \depedge{3}{2}{cc}
      \deproot[edge unit distance=3.75ex]{1}{root}
      \deptrans{Are they right or wrong?}
    \end{dependency}%
    }
    \scalebox{\udscale}{%
    \begin{dependency}
      \tikzstyle{word}=[draw=blue!80!black, shade, top color=blue!40, rounded corners]
      \begin{deptext}[column sep=-0.5mm, row sep=.1ex]
        \color{blue} PRON \& VERB \& CCONJ \& VERB     \& NOUN   \& PUNCT \\
        \color{blue} oni  \& |[word]| Mají \& |[word]| či \& |[word]| nemají \& |[word]| pravdu \& |[word]| ? \\
        \color{blue} they \& have \& or    \& not-have \& truth  \& ?     \\
      \end{deptext}
      \depedge[edge unit distance=3.75ex]{2}{6}{punct}
      \depedge{2}{5}{obj}
      \depedge{2}{4}{conj}
      \depedge{4}{3}{cc}
      \depedge[edge unit distance=4ex,style={blue,thick},label style={draw=blue,thick,fill=white}]{4}{1}{\color{blue}\bf nsubj}
      \depedge[style={blue,thick},label style={draw=blue,thick,fill=white}]{2}{1}{\color{blue}\bf nsubj}
      \depedge[style={blue,thick},label style={draw=blue,thick,fill=white}]{4}{5}{\color{blue}\bf obj}
      \deproot[edge unit distance=5ex]{2}{root}
      \deproot[edge unit distance=5ex,style={blue,thick},label style={draw=blue,thick,fill=none}]{4}{\color{blue}\bf root}
      \deptrans{Are they right or wrong?}
    \end{dependency}%
    }
    \caption{Basic (top) and enhanced (bottom) UD representation of sentence (\ref{ex:majicinemaji}).}
    \label{fig:majicinemajiUD}
  \end{center}
\end{figure}

\subsubsection{Ellipsis}

Due to the adopted restrictions (esp. not allowing extra nodes to be added to
the graph), the representation of ellipis is problematic in both frameworks.
When the governor of a dependent is omitted, it is impossible to construct a
transparent dependency tree.

In PDT, an ``orphan'' dependent receives the
label it would bear if the ellipsis were absent, but with an ellipsis flag
(displayed as the {\tt \_E} suffix in diagrams), explicitly indicating that
the node lacks its governor. The orphan is placed in the position that would
have been occupied by the elided governor (recursively, if necessary, through
further elided governors higher up). See Fig.~\ref{fig:nekterareseni-A},
there are two orphans in the second clause.\footnote{Note that ellipses are
primarily resolved at the t-layer in PDT, allowing the node insertion (see more in \citealp{mikulova-2014-semantic,hajicova-etal-2015-reconstructions}).}

In UD, the promoting strategy is applied: an orphan is “promoted” to the role
of the governor and inherits the governor's dependency label. This strategy
is applied primarily in nominals. E.g., in (\ref{ex:nekterareseni}), the noun
\textit{řešení} ‘solutions’ is elided in the nominal \textit{jiná [řešení]}
‘other [solutions]’ in the second clause. The orphan \textit{jiná} ‘other’
has been promoted to the role of the noun, i.e., to the role of the subject.
However, in the second clause the predicate is also elided, thus the promoted
subject \textit{jiná} ‘other’ is consequently promoted to the head of the
clause and, via a {\tt conj} relation, it is conjoined with the predicate in
the first clause. In such cases (when the elided element is a predicate and
the promoted element is one of its arguments or adjuncts), the {\tt orphan}
relation is used to attach other orphans to the promoted head. This is the
case of the negative particle \textit{nikoli} ‘not’ in
(\ref{ex:nekterareseni}); cf. Fig.~\ref{fig:nekterareseni-UD}.

The PDT and UD frameworks adopt different approaches to representation of
ellipsis. PDT provides more detailed information, in particular through the
explicit marking of ellipsis and the preservation of the proper syntactic
relation. This is a novelty in PDT-C~2.0 compared to earlier releases of the
Prague treebanks, and it enables a more accurate conversion into UD.

\subsection{Dependency Relations}
\label{relation}

The smallest overlap between the frameworks is in the inventory of syntactic
relations. In PDT, syntactic relations are divided into surface syntactic
(annotated at the a-layer, 23 types are distinguished) and deep syntactic
ones (t-layer, over 50 types).

In UD, 37 dependency relations are defined. The relations are specified
syntactically but, unlike in PDT, they are more related to dependent POS
category. E.g., in UD, there are two types of relations for dependents that
functionally correspond to adverbials: the {\tt obl} relation is for
nominal/prepositional phrases (such as \textit{na příkladu} ‘on an-example’
in (\ref{ex:main})) and {\tt advmod} for adverbs (such as \textit{názorně}
‘illustratively’ in \textit{vysvětlit to názorně} ‘to-explain it
illustratively’). In PDT, all adverbials are uniformly labeled as {\tt Adv}
relations (Fig.~\ref{fig:UD-PDT}).

Some additional phenomena, especially deep syntactic ones, e.g., agent and
patient in passives, are handled in UD by means of optional subtypes of the
basic relations. The extent to which the conversion can model such
phenomena is limited by the fact that it primarily relies on information from
the a- and m-layers. E.g., passives are detected by the morphological
annotation of passive participles; if they have a dependent with {\tt
Case=Ins}, it is a possible candidate for the oblique agent, which is
normally expressed as instrumental in Czech.

\subsubsection{Pragmatic Elements}

In both projects, some solutions have to be adopted for the words that are
primarily characterised by pragmatic properties and are not typical
dependents, i.e., words with dominant emphasizing, epistemic, assessment
etc.\ function.

In the PDT framework, these words are POS-tagged as `particles' following the Czech
linguistic tradition.\footnote{This approach is widely used in Slavic and
Central European linguistics, cf.\ \citet{rozumko2016epistemic}.} Although
not understood as parts of surface syntax, they are placed at the a-layer
with the label for pragmatic expression {\tt AuxZ}
\cite{mikulova-etal-2025-form}.

In general UD guidelines, no special attention is paid to these expressions
and they are mainly merged with adverbs ({\tt ADV}). The type of dependency
relation also varies -- depending on the POS category of both the dependent
and the parent, it typically involves the {\tt advmod} dependency relation
(indistinguishable from ordinary adverbials). Such mixing of adverbs and
other expressions may be considered problematic \cite{cecchini2024}.

In UD\_Czech-PDTC, pragmatic expressions are labeled as follows: the category
of POS is {\tt PART}\footnote{The UD guidelines define {\tt PART} as a
function word that does not fit the definitions of other parts of speech,
e.g., the negation particle.} and for the syntactic relation {\tt
advmod:emph} is used (a specific variant of {\tt advmod} signalizing
pragmatic item). See \textit{snad} ‘perhaps’ in (\ref{ex:main}) and
Fig.~\ref{fig:UD-PDT}. This solution is consistent with the general
principles of UD and also captures additional information from PDT.

\begin{table*}[t]
  \centering
    \small
  \setlength{\tabcolsep}{6pt}
  \begin{tabular}{lcccccccccc}
    \toprule
      \multirow{2}[2]{*}{Test Data} & \multicolumn{2}{c}{Training Data}
        & \multirow{2}[2]{*}{UPOS}
        & \multirow{2}[2]{*}{XPOS}
        & \multirow{2}[2]{*}{UFeats}
        & \multirow{2}[2]{*}{AllTags}
        & \multirow{2}[2]{*}{Lemmas}
        & \multirow{2}[2]{*}{UAS}
        & \multirow{2}[2]{*}{LAS} \\\cmidrule{2-3}
      & PDT & PDT-C \\
    \midrule\multirow{2}{*}{PDT-C: Overall}
      & \YES & \NO & 99.01 & 97.96 & 98.57 & 97.71 & 99.35 & 95.36 & 93.97 \\
      & \NO & \YES & 99.27 & 98.45 & 98.92 & 98.26 & 99.57 & 96.20 & 95.06 \\
    \midrule\multirow{2}{*}{PDT-C: Faust}
      & \YES & \NO & 98.91 & 96.14 & 96.51 & 95.76 & 98.12 & 86.26 & 82.74 \\
      & \NO & \YES & 99.50 & 97.09 & 97.46 & 96.88 & 99.21 & 91.61 & 89.84 \\
    \noalign{\kern4pt}\multirow{2}{*}{PDT-C: PCEDT}
      & \YES & \NO & 99.22 & 98.17 & 98.46 & 97.95 & 99.53 & 95.78 & 94.97 \\
      & \NO & \YES & 99.68 & 98.89 & 99.00 & 98.76 & 99.78 & 97.07 & 96.40 \\
    \noalign{\kern4pt}\multirow{2}{*}{PDT-C: PDT}
      & \YES & \NO & 99.04 & 98.19 & 98.81 & 97.96 & 99.42 & 96.18 & 94.89 \\
      & \NO & \YES & 99.03 & 98.25 & 98.86 & 98.02 & 99.48 & 96.28 & 95.04 \\
    \noalign{\kern4pt}\multirow{2}{*}{PDT-C: PDTSC}
      & \YES & \NO & 98.80 & 97.35 & 98.22 & 97.02 & 99.06 & 93.44 & 91.54 \\
      & \NO & \YES & 99.50 & 98.66 & 99.09 & 98.53 & 99.63 & 95.55 & 94.33 \\
    \bottomrule
  \end{tabular}
  \caption{Comparison of UDPipe morphosyntactic performance in percents using
  either whole PDT-C or just PDT as training data, evaluated on whole PDT-C and
  also its four subsets.}
  \label{tab:udpipe_results}
\end{table*}

\subsection{Enhanced Universal Dependencies}
\label{enhanced}

UD also provides guidelines for so-called enhanced graphs. The main reason
for adopting Enhanced UD is to make some of the implicit relations between
words more explicit: especially in coordination or ellipsis constructions,
the dependency path between two content words can be very long or not
indicated at all. Therefore, the enhanced graph includes additional direct
dependencies. Besides additional relations, enhanced graphs may also contain
additional abstract nodes that represent elided material.

Cf. \ Fig.~\ref{fig:majicinemajiUD}: in the basic tree, the
dependency between the predicate \textit{nemají} ‘they-have-not’ and the
shared object \textit{pravdu} ‘truth’ is not marked, whereas in the enhanced
graph it is. There is also an abstract node for elided subject.

In \udpdtc{}, all 6 enhancement types defined by UD guidelines
\cite{nivre-etal-2020-universal} are present. When converting to UD,
structural enhancements are mostly derivable from the annotation on the
a-layer (e.g., the way coordination is represented allows explicit capturing
of a dependent shared by the conjuncts: the shared dependent is captured in the PDT framework as a child of the head of the coordination structure. It does not carry the {\tt \_Co} flag, which signals that it is a shared dependent of the conjuncts, rather than another conjunct; cf.
Fig.~\ref{fig:majicinemajiA}).

\section{Conversion Highlights}
\label{discussion}

A number of factors can be identified when transferring data from one
framework to another. According to our analysis, the following three types
seem to be the most crucial:

\textbf{Framework}. Both frameworks are based on the dependency concept, but
their designs differ for certain phenomena. This may result from different
conventions or from distinct focuses (the design of PDT is based on Czech,
while UD aims for universality). This concerns the treatment of function
words, coordination, and ellipsis (cf. Sect.~\ref{relation}). The differences
here lie in the mode of representation rather than in the interpretation of
the linguistic phenomena themselves, making conversion straightforward in
such cases.

\textbf{Language}. Differences also arise from the language being
converted—in our case, Czech. The language’s specific features must be mapped
onto the universal UD categories in some way. E.g., Czech lacks
articles, and more generally, determiners, but the {\tt DET} category in UD
is useful for distinguishing attributive pronouns (cf. Sect.~\ref{pos}).
These differences are addressed individually during the conversion process,
typically through explicit mapping rules or carefully curated rule lists.

\textbf{Theory}. Different linguistic traditions may influence the difference
in frameworks. For instance, the distinction between adverbs and particles,
common in the Czech linguistic tradition (and therefore also in PDT), is not
universally accepted. However, UD enables such specifics to be captured,
which can be beneficial for processing other languages or for language
research in general.

\section{UDPipe \udpdtcz{} Model}
\label{udpipe}

To quantify the effect of larger and more diverse morphosyntactic data, we
train a UDPipe parser~\citep{straka-etal-2019-udpipe,straka-2018-udpipe}
using both the whole \udpdtcz{} 2.16 train data and also just their PDT
subset. The evaluation of the two parsers on the whole test set and also its
four subsets is presented in Table~\ref{tab:udpipe_results}.

On the whole test set, using the complete training data reduces morphological
tagging errors by 24\%, lemmatization errors by 33\%, and syntactic parsing
errors by 18\%. Given that performance on the PDT subset is improved only
marginally, the error reduction in the other subsets (i.e., in other genres
and domains) is therefore even higher.

We release the model trained on the \udpdtcz{} data under the CC BY-NC-SA
license at {\small\url{http://hdl.handle.net/11234/1-6117}}.

\section{Conclusion}
\label{conclusion}

We have presented the UD conversion of Prague Dependency Treebank --
Consolidated~2.0, currently the largest manually annotated treebank of a
Slavic language, and one of the largest UD treebanks in general. This is a
significant milestone---until release 2.15, UD contained the pilot PDT (1.5M
words), which has now been joined by three other treebanks from the Prague
family (totaling 3.5M words). The expansion also leads to higher genre
diversity (news, popular science, finance, spoken dialogues, user-generated
content).

The annotation scheme of PDT is heavily influenced by Czech and the Czech
linguistic school (e.g.\ the treatment of particles, pronouns vs.\
determiners, ellipsis). Unlike UD, PDT does not aspire to be ``universal'',
i.e., to suit languages typologically different from Czech without
modification.
Nevertheless, the two frameworks share many similarities, and the rich,
multi-layer annotation of PDT provides enough information for most
distinctions required in UD; therefore, the conversion can be performed in
high quality. Optional features, relation subtypes and enhanced UD graphs are
employed to preserve as much from the PDT annotation as possible.

\udpdtc{} has led to more precise and robust parsing models. The UDPipe
parser is currently used for annotating other resources, such as the parallel
texts of InterCorp \cite{rosen-2023-intercorp},
and Old Czech Texts \cite{zeman-etal-2023-old-czech}.

\section*{Limitations}
\label{limitations}

While the core of PDT-C is richly annotated at all three layers (m-layer for
morphology, a-layer for surface syntax and t-layer for deep syntax), about
one fifth of the corpus lacks the t-layer. The conversion procedure thus
cannot rely on t-layer being available (or, alternatively, we could convert
only 80\% of the data to UD). The t-layer could potentially help with certain
phenomena, in particular with reconstruction of elided predicates in Enhanced
UD, but this potential is not exploited at present.

\section*{Acknowledgments}
The research reported here has been supported by the LINDAT/CLARIAH-CZ
Research Infrastructure (\url{https://lindat.cz}), supported by the Ministry
of Education, Youth and Sports of the Czech Republic (Project No. LM2023062).

\section*{Bibliographical References}\label{sec:reference}

\bibliographystyle{lrec2026-natbib}
\bibliography{custom}

\section*{Language Resource References}
\label{lr:ref}

\bibliographystylelanguageresource{lrec2026-natbib}
\bibliographylanguageresource{languageresource}

\section{Appendix}
\label{sec:appendix}

The PDT-to-UD conversion software is open-source, available at
\url{https://github.com/ufal/hamledt} and
\url{https://github.com/ufal/treex}. Its central part is the conversion of
morphological tags and dependency labels. The former is documented at
\url{https://universaldependencies.org/tagset-conversion/cs-pdtc-uposf.html}.
Table~\ref{tab:afun2deprel} gives an idea about the dependency
correspondences, although it must be noted that individual label mappings
often depend on additional conditions and may be accompanied by
transformations of the tree structure.

\begin{table}[ht]
  \begin{tabular}{ll}
    \bf PDT   & \bf UD \\
    \tt Adv   & \tt advmod, obl, advcl \\
    \tt Apos  & \tt appos \\
    \tt Atr   & \tt amod, nummod, det, nmod, \\
              & \tt acl, flat \\
    \tt Atv   & \tt advcl:pred \\
    \tt AuxC  & \tt mark, fixed \\
    \tt AuxG  & \tt punct \\
    \tt AuxK  & \tt punct \\
    \tt AuxO  & \tt discourse \\
    \tt AuxP  & \tt case, fixed \\
    \tt AuxR  & \tt expl:pass \\
    \tt AuxT  & \tt expl:pv \\
    \tt AuxV  & \tt aux, aux:pass \\
    \tt AuxX  & \tt punct \\
    \tt AuxY  & \tt mark, cc, case, advmod, \\
              & \tt compound, fixed \\
    \tt AuxZ  & \tt cc, advmod:emph \\
    \tt Coord & \tt conj \\
    \tt Denom & \tt root, parataxis \\
    \tt Obj   & \tt obj, iobj, obl:arg, ccomp, \\
              & \tt xcomp \\
    \tt Partl & \tt discourse \\
    \tt Pnom  & \tt cop \\
    \tt Pred  & \tt root, parataxis \\
    \tt Sb    & \tt nsubj, nsubj:pass, csubj, \\
              & \tt csubj:pass \\
    \tt Vocat & \tt vocative \\
  \end{tabular}
  \caption{Approximate correspondence of dependency relations.}
  \label{tab:afun2deprel}
\end{table}

\end{document}